\documentclass[letterpaper]{article} 
\usepackage{aaai24}  
\usepackage{times}  
\usepackage{helvet}  
\usepackage{courier}  
\usepackage[hyphens]{url}  
\usepackage{graphicx} 
\usepackage{natbib}  
\usepackage{caption} 
\graphicspath{ {./images/} }

\title{Effect of Adapting to Human Preferences on Trust in Human-Robot Teaming}
\author {
    Shreyas Bhat\textsuperscript{\rm 1},
    Joseph B. Lyons\textsuperscript{\rm 2},
    Cong Shi\textsuperscript{\rm 3},
    X. Jessie Yang\textsuperscript{\rm 1}
}
\affiliations {
    \textsuperscript{\rm 1}University of Michigan\\
    \textsuperscript{\rm 2}Air Force Research Laboratory\\
    \textsuperscript{\rm 3}Miami Herbert Business School\\
    shreyasb@umich.edu, joseph.lyons.6@us.af.mil, congshi@bus.miami.edu, xijyang@umich.edu
}

\usepackage{amsmath}
\usepackage{xcolor}
\usepackage{subcaption}

\begin{document}

\maketitle

\begin{abstract}
We present the effect of adapting to human preferences on trust 
in a human-robot teaming task. The team performs a task in which the robot acts as an action recommender to the human. It is assumed that the behavior of the human and the robot is based on some reward function they try to optimize. We use a new human trust-behavior model that enables the robot to learn and adapt to the human's preferences in real-time during their interaction using Bayesian Inverse Reinforcement Learning.
We present three strategies for the robot to interact with a human: a non-learner strategy, in which the robot assumes that the human's reward function is the same as the robot's, a non-adaptive learner strategy that learns the human's reward function for performance estimation, but still optimizes its own reward function, and an adaptive-learner strategy that learns the human's reward function for performance estimation and also optimizes this learned reward function. Results show that adapting to the human's reward function results in the highest trust in the robot.
\end{abstract}

\section{Introduction}
As autonomous technologies become more ubiquitous, the need to ensure that these technologies behave in a trustworthy manner increases. When working with humans in collaboration, these technologies (e.g., autonomous robots, intelligent decision aids, etc.) are being perceived more as teammates rather than tools to be used by a human operator. In such hybrid teams, trust has been identified as a key factor to facilitate effective and efficient collaboration \cite{sheridan_humanrobot_2016, yang_toward_2023}. To enable such trust-driven partnerships, it is essential for a robot to be able to estimate its human partner's level of trust in real time. Further, it also needs a way to estimate the human's behavior based on her level of trust. Finally, many robotic decision-making systems use reward maximization to plan their behaviors. In such cases, it is necessary to ensure that the ``values" of the robot match that of the human. This is usually accomplished via Inverse Reinforcement Learning \cite{Ng2000}, which aims at learning reward functions through observed behaviors. 

Although it has been theorized that matching the robot's reward function with that of the human in a collaborative task is good for the team, its effect on trust has not been studied in detail. Yet there are two reasons to suggest that such adaptation could be beneficial for trust. First, research has shown that agent adaptation to humans can enhance performance in a HAT context \cite{Chiou2021, Azevedo-Sa_2020}.  Second, agent adaptation could be viewed as the agent being responsive to the human and may, in turn, increase human trust of the agent \cite{Li2021}. This study investigates the effect on trust when humans interact with robots with different interaction strategies. We compare three types of interaction strategy: (1) the robot does not align its reward with the human, (2) the robot does not align its reward with the human, but uses the estimated human reward function for performance assessment, trust estimation, and behavior prediction, and (3) the robot aligns its reward to that of the human. We conduct a human-subjects study with 12 participants. Our results indicate that adapting to human preferences leads to the highest level of trust of the human on the robot, and leads to a significantly higher number of agreements with the robot's recommendations. We use the term ``robot" and ``intelligent agent" interchangeably in this paper.

The rest of the paper is organized as follows: Section \ref{sec:related-work} gives an overview of related work that our study builds upon. Section \ref{sec:formulation} details the human-robot team task and formulates our problem as a trust-aware Markov Decision Process (trust-aware MDP). Section \ref{sec:experiment} details the human-subjects experiment. Section \ref{sec:results} discusses major results and their implications. Finally, section \ref{sec:conclusion} concludes our study and discusses limitations and future work. 

\section{Related Work}
\label{sec:related-work}
This work is motivated by two bodies of research, quantitative trust models and value alignment in HRI.
\subsection{Trust Models in HRI}

\citet{Xu2012} proposed a reputation based trust model to adapt the behavior of robots when trust crossed a certain threshold. Later, the authors proposed an Online Probabilistic Trust Inference Model (OPTIMo) \cite{Xu2015} which modeled trust through a Dynamic Bayesian Network. Subsequently, they used this model to adapt a robot's behavior depending on the trust level of the human \cite{Xu2016}. Our work differs from this in the way that we do not use a threshold-based adaptation strategy. Rather, we use an embedded human-behavior model to predict human behavior with trust and use it directly in the decision-making process of the robot. 

Guo and Yang modeled human trust as a beta distribution with personalized parameters that are updated during interaction depending on the performance of the robot \cite{Guo2021,yang_trust_2023}. They presented simulation results using a reverse-psychology human behavior model and found that the robot can ``manipulate'' human's trust if an explicit trust-gaining reward is not added to its reward function \cite{Guo2021b}. \citet{Bhat2022} used this model with a trust-gaining reward term and demonstrated its usage in a human-subjects study. Our work is similar to this work in the sense that we use the same trust estimation model, but we use a different human behavior model, which we call the ``bounded-rationality-disuse" model, which gets rid of this trust manipulation issue as long as the values of the human and the robot are aligned. Thus, no trust-gaining reward term is needed in the robot's reward function, which makes the reward function much simpler. 

\subsection{Value Alignment in Human-Robot Teams}
The problem of aligning the ``values" of the robot to that of its human teammate has been studied in human-robot teaming literature \cite{Hadfield-Menell2016, Milli2017, Fisac2020, Brown2020, Yuan2022}. In a majority of these works, techniques from Inverse Reinforcement Learning (IRL) \cite{Ng2000} are used to estimate a reward function that aligns with human demonstrations, preferences, or actions. 

A bidirectional value alignment problem is studied in \cite{Yuan2022}. The human knows the true reward function and behaves accordingly while interacting as a supervisor to a group of worker robots. The robots try to learn this true reward function through correctional inputs to their behavior from the human. The human, on the other hand, tries to update her belief on the robot's belief of the true reward function and inputs corrections to their behavior accordingly. In our case, there is no \emph{true} reward function: the human and the robot have their own reward functions, and we want to see the effect of aligning/not aligning the robot's reward function with that of the human.

A ``driver's test" to verify value alignment between the human and the robot is provided in \citet{Brown2020}. This is especially relevant when the human and the robot are performing separate tasks in collaboration, since in this case, it is not enough to match the reward functions of the human and the robot. In our case, since the action sets for the robot and the human are the same, it is enough to match their reward functions to guarantee value alignment. 

We use a Bayesian framework for IRL \cite{Ramachandran2007} which learns human preferences by maintaining and updating a distribution over the possible preferences of the human. The update happens in a Bayesian way after observing the human's selected action. 

\section{Problem Formulation}
\label{sec:formulation}

\subsection{Human-Robot Team Task}
We designed a scenario in which the human-robot team performs a search for potential threats in a town. The team sequentially goes through search sites to look for threats. At each site, the team is given a probability of threat presence inside the site via a scan of the site by a drone. The robot additionally, has some prior information about the probability of threat presence at all of the search sites. This prior information is unknown to the human. After getting the updated probability of threat presence, the robot generates a recommendation for the human. It can either recommend that human use or not use an armored robot for protection from threats. Encountering a threat without protection from the armored robot will result in injury to the human. On the other hand, using the armored robot takes extra time since it takes some time to deploy and move the armored robot to the search site. The goal of the team is to finish the search mission as quickly as possible while also maintaining the soldier's health level. Thus, a two-fold objective arises with conflicting sub-goals: To save time you must take risks, and if you want to avoid risks, you must sacrifice precious mission time. 

\subsection{Trust-Aware Markov Decision Process}
We model the interaction between the human and the robot as a trust-aware Markov Decision Process (trust-aware MDP). A trust-aware MDP is a tuple of the form $(S, A, T, R, H)$, where $S$ is a set of states one of which is the trust of the human in the robot, $A$ is a finite set of actions, $T$ is the transition function giving the transition probabilities from one state to another given an action, $R$ is a reward function and $H$ is an embedded human trust-behavior model, which gives the probabilities of the human choosing a certain action given the action chosen by the robot, their level of trust, etc. 
\subsubsection{States}
The level of trust $t \in [0, 1]$.
\subsubsection{Actions}
The recommender robot has two choices of action: recommend to use or not use the armored robot. These are represented by $a^r=1$ and $a^r=0$ respectively. 
\subsubsection{Reward Function}
The rewards for both agents (the human and the robot) are a weighted sum of the negative cost of losing health and losing time. The weights for these costs can be different for the robot and the human. For agent $o\in\{h,r\}$, the reward function can be written as,

\begin{equation}
    \label{eq:reward-function}
    R^o(D, a) = -w^o_h h(D, a) - w^o_c c(a).
\end{equation}
Here, $D$ is a random variable representing the presence of threat inside a search site, $a$ is the action chosen by the human to implement, $o\in\{h,r\}$ represents the agent, either the human $h$ or the robot $r$. $h(D, a)$ gives the health loss cost and $c(a)$ gives the time loss cost. 

\subsubsection{Transition Function}
The transition function gives the dynamics of trust as the human interacts with the robot. We use the model from \cite{Guo2021, Bhat2022} which models trust as a random variable following the Beta distribution based on personalized parameters $(\alpha_0, \beta_0, w^s, w^f)$. 

\begin{align}
\begin{split}
    \label{eq:trust-update}
    t_i &\sim Beta(\alpha_i, \beta_i),\\
    \alpha_i &= \alpha_0 + \sum_{j=1}^ip_j,\\
    \beta_i &= \beta_0 + \sum_{j=1}^i(1-p_j).
\end{split}
\end{align}
Where $i$ is the number of interactions completed between the human and the robot, $t_i$ is the current level of trust, and $p_j$ is the realization of the random variable performance $(P_j)$ of the recommender robot at the $j^{th}$ interaction,

\begin{equation}
\label{eq:performance}
    P_j =
    \begin{cases}
        1, ~\text{ if } R^h_j(a^r_j) \geq R^h_j(1-a^r_j),\\
        0, ~\text{ otherwise.}
    \end{cases}
\end{equation}
Here, $R^h_j(a^r_j)$ is the reward for the human for choosing the recommended action $(a^r_j)$ at the $j^{th}$ interaction and $R^h(1-a^r_j)$ is the same for the other action. 

\subsubsection{Human Trust-Behavior Model}
\label{sec:behavior-model}
A human trust-behavior model gives the probabilities of a human choosing an action, given the robot's action, their trust level, and other factors such as the human's preferences. In our study, we use the \emph{Bounded Rationality Disuse Model} as the human trust-behavior model. This model states that the human chooses the recommended action with a probability equal to the human's current level of trust. If the human chooses to ignore the recommendation, she will choose an action according to the bounded rationality model of human behavior. That is, she will choose an action with a probability that is proportional to the exponential of the expected reward associated with that action. Mathematically,

\begin{align}
\label{eq:behavior-model}
    P(a^h_i=a|a^r_i=a) &= t_i + (1-t_i)q_a,\\
    P(a^h_i=1-a|a^r_i=a) &= (1-t_i)(1-q_a).
\end{align}
where $q_a$ is the probability of choosing action $a$ under the bounded rationality model,

\begin{equation}
    q_a = \frac{\exp(\kappa E[R^h_i(a)])}{\sum_{a'\in\{0,1\}}\exp(\kappa E[R^h_i(a')])}
\end{equation}

\subsection{Bayesian Inverse Reinforcement Learning}
We use Bayesian IRL to estimate the reward weights of the human as they interact with the recommender robot. This is accomplished by maintaining a distribution of the possible reward weights and updating it using Bayes' rule after observing the human's behavior. More precisely, if $b_i(w)$ is the belief distribution on the reward weights before the $i^{th}$ interaction, the distribution after the $i^{th}$ interaction is given by,

\begin{equation}
\label{eq:Bayes-IRL}
    b_{i+1}(w) \propto 
    \begin{cases}
        P(a^h_i=a^r_i|a^r_i)b_i(w), \phantom{1-a}\text{ if } a^h_i = a^r_i,\\
        P(a^h_i=1-a^r_i|a^r_i)b_i(w), ~\text{ otherwise.}
    \end{cases}
\end{equation}

In our formulation, we only learn a distribution over the health reward weight of the human, $w^h_h$, and assume that the time reward weight is defined by $w_c^h := 1-w^h_h$. Further, we use the mean of the maintained distribution as an estimate for the human's health reward weight. 

\section{Experiment}
\label{sec:experiment}
This section provides details about the testbed used for data collection and the human-subject experiment. The experiment complied with the American Psychological Association code of ethics and was approved by the Institutional Review Board at the University of Michigan.
\subsection{Testbed}
\begin{figure}[h]
    \centering
    \includegraphics[width=0.9\linewidth]{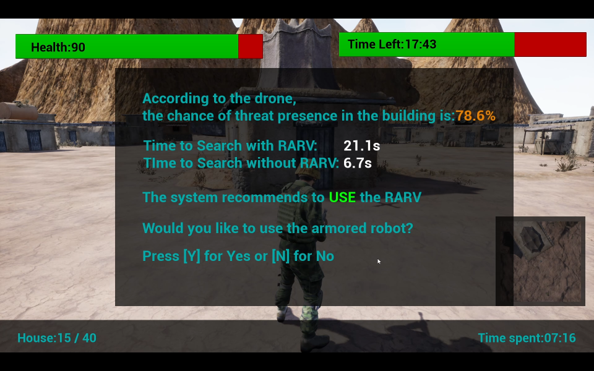}
    \caption{The recommendation interface}
    \label{fig:rec-interface}
\end{figure}

We updated some elements of the testbed from \cite{Bhat2022} for this study, following feedback from participants in that study. The testbed was developed in the Unreal Engine game development platform. We updated the recommendation interface to be more informative and to help the participants make their own decisions if they choose to do so. In the updated interface (shown in fig. \ref{fig:rec-interface}), the participants are shown the probability of threat presence reported by the drone, the recommendation given by the intelligent agent, and an estimate of search time with and without the armored robot. In order to better separate the threat detection task from the recommendation task, the participants were told that a separate entity called an \emph{intelligent agent} will given them action recommendations. On the other hand, the drone's task is to just scan a site and report the threat level inside it. The participants were specifically asked to report their trust level on the intelligent agent.

\begin{figure}[h]
    \centering
    \includegraphics[width=0.9\linewidth]{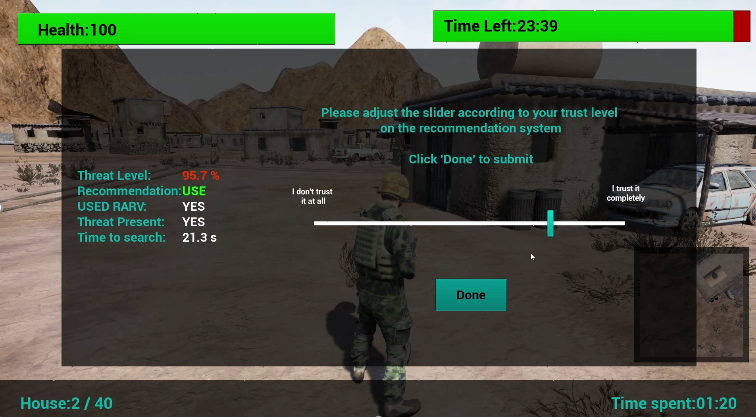}
    \caption{The trust feedback slider used to get feedback from the participants after every search site. The mission timer is paused when the slider is shown to let the participants take their time in adjusting their trust.}
    \label{fig:feedback-slider}
\end{figure}

Further, we updated the trust feedback slider to provide information about the last interaction that the participant had, in order to help them make an informed decision about their trust. The updated interface can be seen in fig. \ref{fig:feedback-slider}. 

We designed a within-subjects study. Our goal is to compare trust between different interaction strategies. Given the high variation between trust dynamics between participants \cite{Bhat2022}, we think that it is better to compare trust within-subjects than between-subjects. The participants completed 3 missions in total. In each mission, they interacted with an intelligent agent following one of the 3 interaction strategies (detailed in sec. \ref{sec:strategies}). In each mission, they sequentially searched through 40 search sites. The condition order was counterbalanced using a Latin square.

\subsection{Participants}
We collected data from $12$ participants (Age: Mean = $21.92$ years, \textit{SD} = $2.36$). All participants were students from the College of Engineering at the University of Michigan. 
\subsection{Interaction Strategies}
\label{sec:strategies}
We designed three interaction strategies for the intelligent agent:
\begin{itemize}
    \item \textbf{Non-learner:} The intelligent agent does not learn the reward weights of the human. It assumes that the human and the intelligent agent share the same reward weights.
    \item \textbf{Non-adaptive learner:} The intelligent agent learns personalized reward weights for each human. It only uses these learned weights for performance assessment and human behavior modeling. It still optimizes the MDP based on its own fixed reward weights. 
    \item \textbf{Adaptive learner:} The intelligent agent learns personalized reward weights for each human. It uses them for performance assessment, human behavior modeling, and also optimizes the MDP based on these reward weights. In other words, it updates its own reward function match the learned reward function. 
\end{itemize}
Although it may look like the non-learner and the non-adaptive learner both optimize the same reward function, they actually optimize expected reward under the assumed human trust-behavior model. Thus, since the non-adaptive-learner has a better estimate of the human's preferences, we postulate that it will have a better estimate of the human's trust and behavior, and hence, will show some difference in its recommendations compared to the non-learner. 

\subsection{Measures}
\subsubsection{Pre-experiment Measures}
Before the beginning of each mission, we ask the participants to rate their preference between saving the soldier's health and saving time by moving a slider between these two objectives, showing their relative importance.

\subsubsection{In Experiment Measures}
After each site's search was completed, the participants were asked to provide feedback on their level of trust on the intelligent agent's recommendations. The interface can be seen in fig. \ref{fig:feedback-slider}. The slider values were between 0 and 100 with a step of 2 points. The end-of-mission trust (used in sec \ref{sec:results}) is the feedback given by the participant using this slider after completing the search of the last search site.
\subsubsection{Post-mission Measures}
We used the following measures as a post-mission survey that the participants filled out after every mission. 
\begin{itemize}
    \item \textbf{Post-mission Trust:} This was measured using Muir's trust questionnaire \cite{Muir1996}. It has 9 questions each with a slider answer range between 0 and 100. Note that this is separate from the end-of-mission trust, which is a subjective rating on a single slider by the participant after the last search site is completed.
    \item \textbf{Post-mission Reliance Intentions:} Measured using the scale developed in \cite{Lyons2019}. 
    It has 10 items but 6 items were used herein, each on a 7-point Likert scale.
\end{itemize}
\section{Results and Discussion}
\label{sec:results}
This section provides an overview of our major results and discusses some reasons behind them and their implications. Note: All error bars on bar charts are standard errors.
\subsection{Trust}
We expect that adapting to human preferences will result in higher levels of trust of the human on the robot. 

\begin{figure}[h]
    \centering
    \includegraphics[width=0.8\columnwidth]{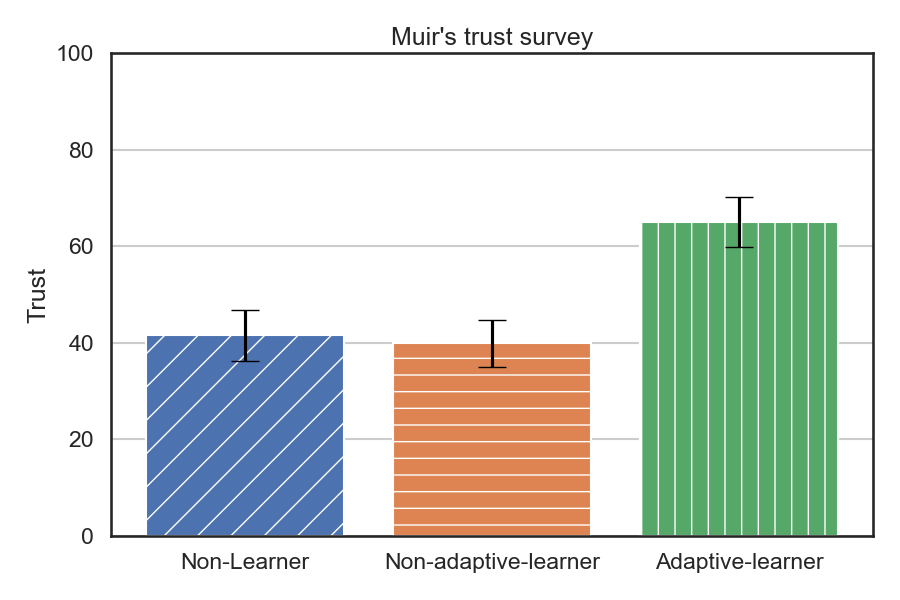}
    \caption{Mean and standard error (SD) of trust ratings given by the participants post-mission using Muir's trust scale \cite{Muir1996}}
    \label{fig:trust-muir}
\end{figure}

Fig. \ref{fig:trust-muir} shows the post-mission trust rating given using Muir's trust scale. Repeated measures ANOVA shows significant difference between the trust ratings given to the three strategies ($F(2,22) = 11.962, p < 0.001$), with the highest trust given to the adaptive. Post-hoc analysis with Bonferroni adjustment shows significant differences between the non-learner strategy and the adaptive-learner strategy ($p=0.001$) and between the non-adaptive-learner strategy and the adaptive-learner strategy ($p=0.014$).

\begin{figure}[h]
    \centering
    \includegraphics[width=0.8\columnwidth]{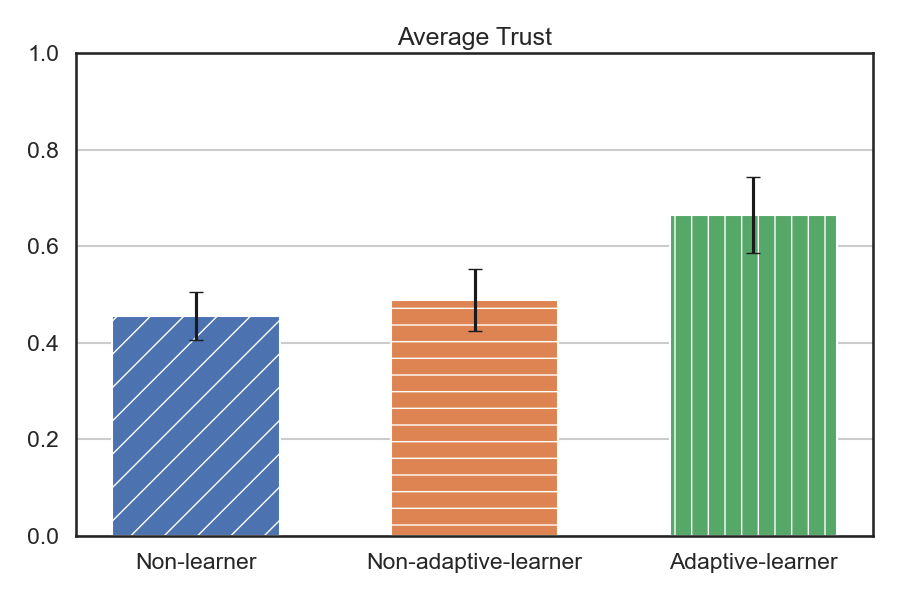}
    \caption{Average trust reported by the participants across the interaction period, mean and standard error.}
    \label{fig:average-trust}
\end{figure}

Fig. \ref{fig:average-trust} shows the average trust rating given by the participants to the recommendations of the intelligent agent across their interaction period. Repeated measures ANOVA shows significant difference between the three strategies $(F(2, 22) = 4.968, p = 0.017)$. Post-hoc analysis with Bonferroni adjustment reveals significant difference in average trust rating between the non-learner and the adaptive learner strategy $(p = 0.044)$ and no significant difference between the other two pairs. 

\begin{figure}[h]
    \centering
    \includegraphics[width=0.8\columnwidth]{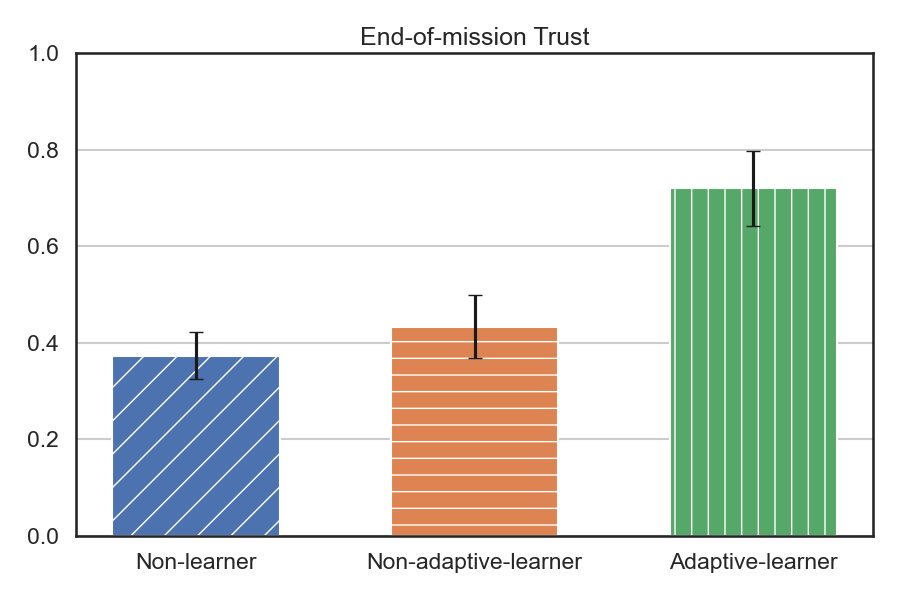}
    \caption{Trust reported by the participants at the end of the mission, mean and standard error.}
    \label{fig:end-trust}
\end{figure}

Fig. \ref{fig:end-trust} shows the trust rating given by the participants to the recommendations of the intelligent agent at the end of their mission. Repeated measures ANOVA shows significant difference between the three strategies $(F(2, 22) = 7.455, p = 0.003)$. Post-hoc analysis with Bonferroni adjustment reveals significant difference in average trust rating between the non-learner and the adaptive learner strategy $(p = 0.044)$ and a marginally significant difference between the non-adaptive-learner and adaptive-learner strategies $(p=0.057)$. This trend could reach significance with more data which we are currently working on. The end-of-mission trust rating should be a stable trust rating since the participants have had enough interactions with the intelligent agent to have a good sense of their trust on it. 

\begin{figure}
    \centering
    \includegraphics[width=0.8\columnwidth]{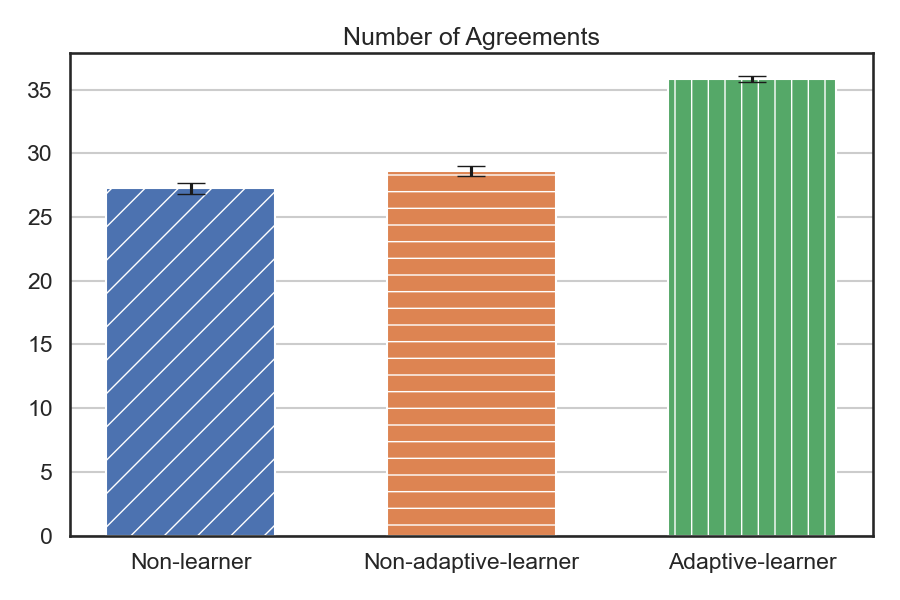}
    \caption{Number of agreements between the recommendation from the robot and the human's action choice, mean and standard error.}
    \label{fig:agreements}
\end{figure}

Fig. \ref{fig:agreements} shows the number of agreements between the recommendation from the intelligent agent and the participant's action selection. We expect there to be a positive correlation between the number of agreements and trust reported by the participants. Repeated measures ANOVA shows significant difference between the three strategies $(F(2, 22) = 13.732, p < 0.001)$. Post-hoc analysis with Bonferroni adjustment reveals significant difference in average trust rating between the non-learner and the adaptive learner strategy $(p = 0.003)$ and between the non-adaptive-learner and the adaptive-learner strategy $(p=0.009)$. The participants agreed the most with the adaptive-learner's recommendations.

\section{Conclusions}
\label{sec:conclusion}
In this study, we provided a demonstration of the use of Bayesian IRL coupled with the bounded-rationality-disuse model of human behavior to learn a human's preferences in performing a human-robot team task. We implemented an adaptive interaction strategy for the robot that learns and optimizes a reward function based on these preferences. We showed the trust and performance improvement when using such an adaptive interaction strategy compared to two baselines. 

The results of our study should be seen in light of the following limitations. First, we provide a demonstration in the case when there are only two components in the team's reward function. Therefore, we only need to learn the human's preference for one of the two components and can ascertain their relative preference between the two objectives. Our formulation, however, can readily be extended to the case where there are more than two objectives in the team's reward function, with additional computations required to learn and maintain a distribution over each reward weight. Second, we used an uninformed uniform prior for the reward weights of the human. This uniform distribution was, thus, also used to set the reward weights for the non-adaptive interaction strategies. This simulates a scenario where we do not have any other way of setting the reward weights for the robot. Another approach could be to use a data-driven prior on the reward weight distribution to set these weights. A similar comparison between the three strategies with such an informed prior could be a direction for a future study. 

\section*{Acknowledgments}

This work was supported by the Air Force Office of Scientific Research under grant \#FA9550-20-1-0406.

\bibliography{aaai24}

\begin{thebibliography}{21}
\providecommand{\natexlab}[1]{#1}

\bibitem[{Azevedo-Sa et~al.(2020)Azevedo-Sa, Jayaraman, Yang, Robert, and
  Tilbury}]{Azevedo-Sa_2020}
Azevedo-Sa, H.; Jayaraman, S.~K.; Yang, X.~J.; Robert, L.~P.; and Tilbury,
  D.~M. 2020.
\newblock Context-Adaptive Management of Drivers’ Trust in Automated
  Vehicles.
\newblock \emph{IEEE Robotics and Automation Letters}, 5(4): 6908--6915.

\bibitem[{Bhat et~al.(2022)Bhat, Lyons, Shi, and Yang}]{Bhat2022}
Bhat, S.; Lyons, J.~B.; Shi, C.; and Yang, X.~J. 2022.
\newblock Clustering Trust Dynamics in a Human-Robot Sequential Decision-Making
  Task.
\newblock \emph{IEEE Robotics and Automation Letters}, 7(4): 8815--8822.

\bibitem[{Brown, Schneider, and Niekum(2020)}]{Brown2020}
Brown, D.~S.; Schneider, J.; and Niekum, S. 2020.
\newblock Value Alignment Verification.
\newblock \emph{CoRR}, abs/2012.01557.

\bibitem[{Chiou and Lee(2023)}]{Chiou2021}
Chiou, E.~K.; and Lee, J.~D. 2023.
\newblock Trusting Automation: Designing for Responsivity and Resilience.
\newblock \emph{Human Factors}, 65(1): 137--165.
\newblock PMID: 33906505.

\bibitem[{Fisac et~al.(2020)Fisac, Gates, Hamrick, Liu, Hadfield-Menell,
  Palaniappan, Malik, Sastry, Griffiths, and Dragan}]{Fisac2020}
Fisac, J.~F.; Gates, M.~A.; Hamrick, J.~B.; Liu, C.; Hadfield-Menell, D.;
  Palaniappan, M.; Malik, D.; Sastry, S.~S.; Griffiths, T.~L.; and Dragan,
  A.~D. 2020.
\newblock Pragmatic-Pedagogic Value Alignment.
\newblock In Amato, N.~M.; Hager, G.; Thomas, S.; and Torres-Torriti, M., eds.,
  \emph{Robotics Research}, 49--57. Cham: Springer International Publishing.
\newblock ISBN 978-3-030-28619-4.

\bibitem[{Guo, Shi, and Yang(2021)}]{Guo2021b}
Guo, Y.; Shi, C.; and Yang, X.~J. 2021.
\newblock Reverse Psychology in Trust-Aware Human-Robot Interaction.
\newblock \emph{IEEE Robotics and Automation Letters}, 6(3): 4851--4858.

\bibitem[{Guo and Yang(2021)}]{Guo2021}
Guo, Y.; and Yang, X.~J. 2021.
\newblock Modeling and Predicting Trust Dynamics in Human-Robot Teaming: A
  Bayesian Inference Approach.
\newblock \emph{International Journal of Social Robotics}.

\bibitem[{Hadfield-Menell et~al.(2016)Hadfield-Menell, Dragan, Abbeel, and
  Russell}]{Hadfield-Menell2016}
Hadfield-Menell, D.; Dragan, A.; Abbeel, P.; and Russell, S. 2016.
\newblock Cooperative Inverse Reinforcement Learning.

\bibitem[{Li et~al.(2021)Li, Ni, Agrawal, Jia, Raja, Gui, Hughes, Lewis, and
  Sycara}]{Li2021}
Li, H.; Ni, T.; Agrawal, S.; Jia, F.; Raja, S.; Gui, Y.; Hughes, D.; Lewis, M.;
  and Sycara, K. 2021.
\newblock Individualized Mutual Adaptation in Human-Agent Teams.
\newblock \emph{IEEE Transactions on Human-Machine Systems}, 51(6): 706--714.

\bibitem[{Lyons and Guznov(2019)}]{Lyons2019}
Lyons, J.~B.; and Guznov, S.~Y. 2019.
\newblock Individual differences in human–machine trust: A multi-study look
  at the perfect automation schema.
\newblock \emph{Theoretical Issues in Ergonomics Science}, 20(4): 440--458.

\bibitem[{Milli et~al.(2017)Milli, Hadfield-Menell, Dragan, and
  Russell}]{Milli2017}
Milli, S.; Hadfield-Menell, D.; Dragan, A.; and Russell, S. 2017.
\newblock Should {Robots} be {Obedient}?
\newblock ArXiv:1705.09990 [cs].

\bibitem[{Muir and Moray(1996)}]{Muir1996}
Muir, B.; and Moray, N. 1996.
\newblock Trust in automation. Part II. Experimental studies of trust and human
  intervention in a process control simulation.
\newblock \emph{Ergonomics}, 39: 429--60.

\bibitem[{Ng and Russell(2000)}]{Ng2000}
Ng, A.~Y.; and Russell, S. 2000.
\newblock Algorithms for Inverse Reinforcement Learning.
\newblock In \emph{in Proc. 17th International Conf. on Machine Learning},
  663--670. Morgan Kaufmann.

\bibitem[{Ramachandran and Amir(2007)}]{Ramachandran2007}
Ramachandran, D.; and Amir, E. 2007.
\newblock Bayesian Inverse Reinforcement Learning.
\newblock In \emph{Proceedings of the 20th International Joint Conference on
  Artifical Intelligence}, IJCAI'07, 2586–2591. San Francisco, CA, USA:
  Morgan Kaufmann Publishers Inc.

\bibitem[{Sheridan(2016)}]{sheridan_humanrobot_2016}
Sheridan, T.~B. 2016.
\newblock Human–{Robot} {Interaction}: {Status} and {Challenges}.
\newblock \emph{Human Factors}, 58(4): 525--532.
\newblock Publisher: SAGE Publications Inc.

\bibitem[{Xu and Dudek(2012)}]{Xu2012}
Xu, A.; and Dudek, G. 2012.
\newblock Trust-driven interactive visual navigation for autonomous robots.
\newblock In \emph{2012 IEEE International Conference on Robotics and
  Automation}, 3922--3929.

\bibitem[{Xu and Dudek(2015)}]{Xu2015}
Xu, A.; and Dudek, G. 2015.
\newblock OPTIMo: Online Probabilistic Trust Inference Model for Asymmetric
  Human-Robot Collaborations.
\newblock In \emph{2015 10th ACM/IEEE International Conference on Human-Robot
  Interaction (HRI)}, 221--228.

\bibitem[{Xu and Dudek(2016)}]{Xu2016}
Xu, A.; and Dudek, G. 2016.
\newblock Maintaining efficient collaboration with trust-seeking robots.
\newblock In \emph{2016 IEEE/RSJ International Conference on Intelligent Robots
  and Systems (IROS)}, 3312--3319.

\bibitem[{Yang, Guo, and Schemanske(2023)}]{yang_trust_2023}
Yang, X.~J.; Guo, Y.; and Schemanske, C. 2023.
\newblock From {Trust} to {Trust} {Dynamics}: {Combining} {Empirical}
  and {Computational} {Approaches} to {Model} and {Predict} {Trust}
  {Dynamics} {In} {Human}-{Autonomy} {Interaction}.
\newblock In Duffy, V.~G.; Landry, S.~J.; Lee, J.~D.; and Stanton, N., eds.,
  \emph{Human-{Automation} {Interaction}: {Transportation}}, Automation,
  {Collaboration}, \& {E}-{Services}, 253--265. Cham: Springer International
  Publishing.

\bibitem[{Yang, Schemanske, and Searle(2023)}]{yang_toward_2023}
Yang, X.~J.; Schemanske, C.; and Searle, C. 2023.
\newblock Toward {Quantifying} {Trust} {Dynamics}: {How} {People} {Adjust}
  {Their} {Trust} {After} {Moment}-to-{Moment} {Interaction} {With}
  {Automation}.
\newblock \emph{Human Factors}, 65(5): 862--878.

\bibitem[{Yuan et~al.(2022)Yuan, Gao, Zheng, Edmonds, Wu, Rossano, Lu, Zhu, and
  Zhu}]{Yuan2022}
Yuan, L.; Gao, X.; Zheng, Z.; Edmonds, M.; Wu, Y.~N.; Rossano, F.; Lu, H.; Zhu,
  Y.; and Zhu, S.-C. 2022.
\newblock In situ bidirectional human-robot value alignment.
\newblock \emph{Science Robotics}, 7(68): eabm4183.

\end{thebibliography}
\end{document}